\documentclass[10pt,twocolumn,letterpaper]{article}
\usepackage{iccv}
\usepackage{graphicx}
\usepackage{comment}
\usepackage{amsmath,amssymb} 
\usepackage{color}
\usepackage{bbm}
\usepackage{caption}
\usepackage{subcaption}
\usepackage{times}

\newcommand{\ud}{\,\mathrm{d}}

\newcommand{\R}{\mathbb{R}}

\newcommand{\cut}[1]{}

\DeclareMathOperator*{\argmin}{arg\,min}

\DeclareMathOperator*{\median}{median}

\usepackage[ruled]{algorithm}
\usepackage{algorithmicx}
\usepackage{algpseudocode}
\usepackage{multirow}
\usepackage{makecell} 
\usepackage[pagebackref=true,breaklinks=true,letterpaper=true,colorlinks,bookmarks=false]{hyperref}

\iccvfinalcopy

\usepackage[breaklinks=true,bookmarks=false]{hyperref}

\def\iccvPaperID{8482} 

\ificcvfinal\fi
\begin{document}

\title{Flow-Guided Video Inpainting with Scene Templates}

\author{Dong Lao \,\,\, Peihao Zhu \,\,\, Peter Wonka \,\,\, Ganesh Sundaramoorthi\\
KAUST, Saudi Arabia\\
{\tt\small \{dong.lao, peihao.zhu, peter.wonka, ganesh.sundaramoorthi\}@kaust.edu.sa}
}

\maketitle
\thispagestyle{empty}

\begin{abstract}
We consider the problem of filling in missing spatio-temporal regions of a video. We provide a novel flow-based solution by introducing a generative model of images in relation to the \emph{scene} (without missing regions) and mappings from the scene to images. We use the model to jointly infer the \emph{scene template}, a 2D representation of the scene, and the mappings. This ensures consistency of the frame-to-frame flows generated to the underlying scene, reducing geometric distortions in flow based inpainting. The template is mapped to the missing regions in the video by a new ($L^2$-$L^1$) interpolation scheme, creating crisp inpaintings and reducing common blur and distortion artifacts. We show on two benchmark datasets that our approach out-performs state-of-the-art quantitatively and in user studies.\footnote{Dataset and code: \href{https://github.com/donglao/videoinpainting}{https://github.com/donglao/videoinpainting} }
\end{abstract}

\section{Introduction}
Video inpainting is the problem of filling spatial-temporal regions, i.e., masked regions, with content that naturally blends with the remaining parts of the video.  This is useful in video editing tasks, including removing watermarks or unwanted objects and video restoration. As videos exhibit temporal regularity, to inpaint a given frame, it is natural to use data from other frames, as the data in other frames may correspond to parts of the scene behind the masked region. Many state-of-the-art methods for video inpainting are flow-guided \cite{huang2016temporally,xu2019deep,oh2019onion,Gao-ECCV-FGVC}, which take the approach of copying unmasked data from other frames into the masked region of a given frame by using optical flow. 

\begin{figure}[t]
\begin{center}
\includegraphics[width=.47\textwidth]{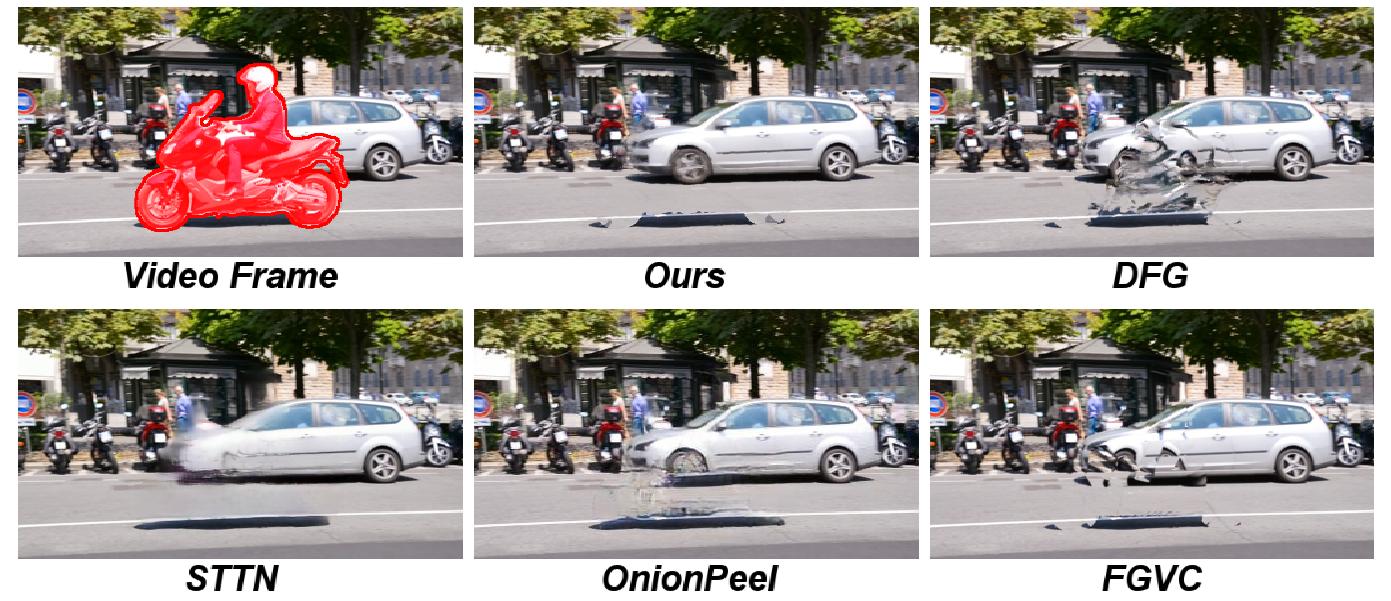}
\end{center}
\caption{{\bf Comparison with state-of-the-art.} Our approach uses a generative model of formation of images from the scene to infer flows that are consistent with the scene. This reduces visual distortions in inpainting compared with state-of-the-art: DFG \cite{xu2019deep}, STTN \cite{zeng2020learning}, OnionPeel \cite{oh2019onion}, FGVC \cite{Gao-ECCV-FGVC}. {\it Animation in the supplementary materials.}}
\label{fig:teaser}
\end{figure}

While these approaches inpaint with plausible data from the scene through other frames, unlike single image inpainting methods (e.g. \cite{efros2001image,yu2018generative}) that attempt to halluncinate image values in the masked region from other regions in the image or learned through datasets, they are highly dependent on the quality of the optical flow. Even though optical flow has advanced significantly with the progress of deep learning to the point that recent methods on benchmark optical flow datasets produce only small errors, there are two complications in applying optical flow to inpainting. First, the flow that is needed in the masked region for inpainting is the flow had the contents within the masked region been removed to reveal the part of the scene that the mask occludes. As this is not possible to determine directly, it is hallucinated, typically learned through data \cite{xu2019deep,Gao-ECCV-FGVC}. However, there is no guarantee that this is consistent with the scene or hallucinations from other frames, producing visual distortions in inpainting. Secondly, even small errors in optical flow can produce noticeable visual distortions in the inpainted video, which is further amplified as flow is aggregated over multiple frames as data from far away frames may be needed. Attempts have been made to reduce these errors by applying temporal regularity to the flow \cite{xu2019deep,zhang2019internal,zeng2020learning,Gao-ECCV-FGVC}, but these naive regularizers (flow between frames is close) may not be consistent with the scene geometry and can still produce visual distortions (Figure \ref{fig:teaser}).


\cut{
Earlier works \cite{wexler2004space,newson2014video,huang2016temporally} rely on patches to reduce the computational complexity of flow estimation but the inpainting results suffer since the coarse patch representation cannot handle complex motion patterns. Recent progress in deep learning enables neural networks to learn both optical flow estimation and video content propagation \cite{xu2019deep,lee2019copy,Gao-ECCV-FGVC}. End-to-end learning methods \cite{kim2019deep,wang2019video,oh2019onion,zhang2019internal,chang2019free,zeng2020learning} further implicitly model this process in their loss functions, and results are highly dependent on the quality and quantity of data. Further, limited by computational power, they either work on only low resolution or process a small batch of frames.
}

In this paper, we aim to reduce visual distortions in video inpainting due to physically implausible and temporally inconsistent flows by deriving a generative model that closely models the physical image formation process in generating images from the scene, and using it to infer flow. This model represents the 3D scene (all of the scene corresponding to the video outside the masked regions) as a 2D \emph{scene template}. The inpainted image at frame $t$ is a mapping (general piecewise smooth warping) of the part of the scene template in view in frame $t$ to the image domain. The model constrains the warps to be consistent with the \emph{scene} and the images. This induces temporal consistency by naturally and efficiently enforcing that all pairwise mappings between images generated from the model must correctly match images. This reduces implausible flow hallucinations common in current approaches. Our inference procedure computes the warps and the scene template jointly from this model, which reduces distortions in inpainting. Our contributions are specifically:\newline
{\bf 1.} We solve flow-guided inpainting by applying a generative model of the scene to images, and using it to infer flow and a model of the scene (the \emph{scene template}). This gives
more temporally consistent and plausible pair-wise flows compared with existing \cut{flow-based inpainting }methods, which results in 
inpainting results that are more temporally consistent and have less geometric distortions. {\bf 2.} We propose a novel $L^2$-$L^1$ combined optimization procedure that generates the inpainting from the scene template together with a interpolation strategy that significantly improves inpainting quality and further reduces geometric distortions and blurring artifacts. {\bf 3}. We introduce the first benchmark dataset ({\it Foreground Removal}) on removing occluding objects from video. We introduce a quantitative protocol while previous art relies on visual comparisons. {\bf 4.} We demonstrate the advantage of our algorithm on the DAVIS \cite{Perazzi2016} and Foreground Removal datasets and demonstrate superior results (both through user studies and quantitatively) compared to state-of-the-art.

\section{Related Work} 
{\bf Video Inpainting:} 
Single image inpainting methods \cite{efros1999texture,efros2001image,yu2019free,yu2018generative,IizukaSIGGRAPH2017,liu2018image} have had success in the past decades. However, when applied to video data, they generally produce artifacts due to a lack of temporal consistency. Early video inpainting methods \cite{wexler2004space,patwardhan2005video,newson2014video} extend patch-based single image techniques to video data. More recent works \cite{huang2016temporally,strobel2014flow,xu2019deep,lee2019copy,zhang2019internal,Gao-ECCV-FGVC} use optical flow or variants to model spatio-temporal correspondence across frames. To hallucinate flow inside the masked region, non-learning approaches \cite{strobel2014flow,huang2016temporally} rely on energy minimization assuming smoothness of the flow; \cite{xu2019deep, Gao-ECCV-FGVC} is a deep learning solution that first computes flow between image pairs, then uses a neural network to hallucinate flow inside the masked region. End-to-end learning methods \cite{kim2019deep,wang2019video,oh2019onion,zhang2019internal,chang2019free,zeng2020learning,li2020short} model cross-frame correspondence in their loss functions. For example, \cite{zhang2019internal} jointly infers appearance and flow while penalizing temporal inconsistency. These methods only process a small fixed number of frames or run at low resolution as they are limited by hardware constraints. As our method does not have such a limitation and produces consistency with the scene, we out-perform these methods.

{\bf Layered Approaches:}
Our method relates to layered approaches \cite{sun2012layered,wang1994representing,lao2018extending,jackson2008dynamic}, which represent a scene as moving 2D layers that can occlude each other. Layered approaches are powerful tools that can be applied to motion segmentation \cite{brox2010object,yang2013modeling,taylor2015causal,yang2015self,lao2017minimum,yang2021dystab,yang2019unsupervised} as they provide a principled way of occlusion reasoning in videos. We adopt a layered formulation to create our scene template using modern advances in optical flow and deep learning, which we then use for inpainting.

\cut{
Our method is motivated by layered approaches \cite{sun2012layered,wang1994representing,lao2018extending,jackson2008dynamic}, which model the video formation process by a series of 2D representations and transformations. Layered approaches naturally provide occlusion reasoning in videos, therefore can be adopted by inpainting that aims to remove the occlusion (masks). In this paper, we use such a layered formulation to model the scene template and warps.
}

\cut{In this paper, we explicitly model the video by a scene template and corresponding warps between the template and images. This is motivated by layered approaches, which decompose image sequences into layers by different motion. A layered formulation provides a principled way of occlusion reasoning in videos, thus handles both moving objects and moving backgrounds. Specifically, we model a background layer that serves as the scene template.}

\section{Computing the Scene Template}

\begin{figure*}[t]
\centering
\includegraphics[width=0.8\textwidth]{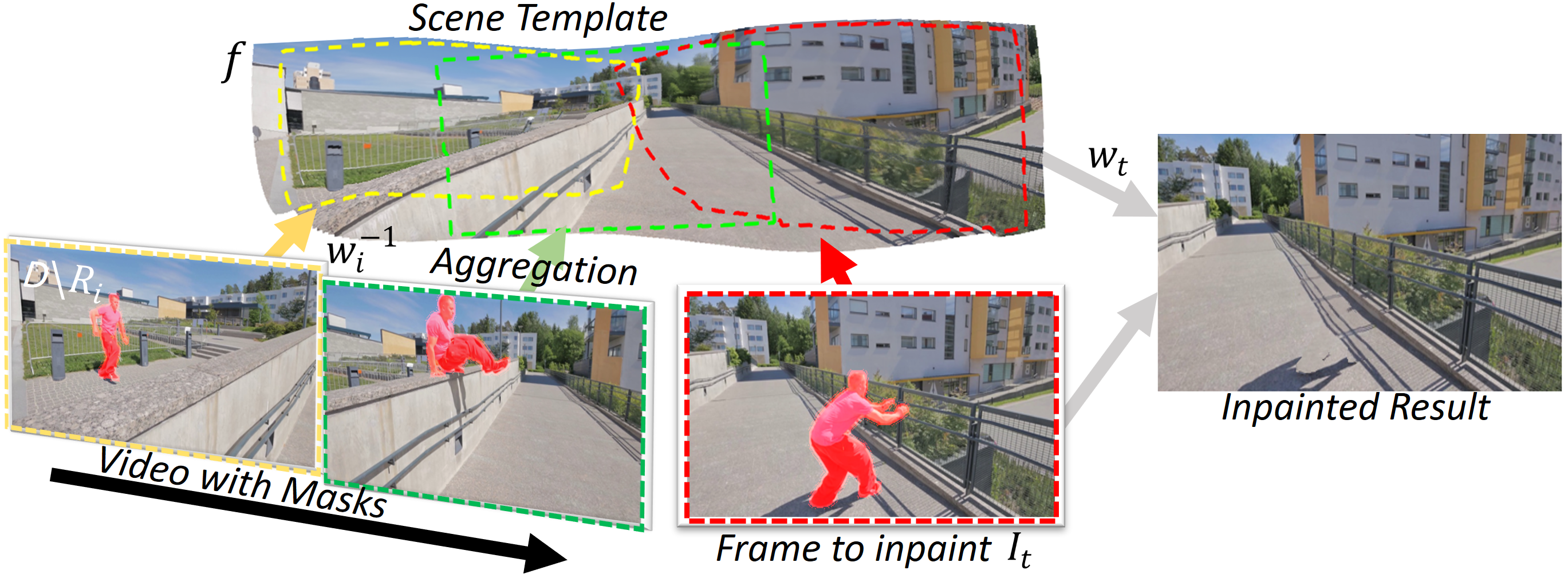}
  \caption{{\bf Schematic of our approach.} Unmasked regions in images are mapped via piece-wise smooth warps $w_i^{-1}$ to create the scene template. Note portions of the scene template that correspond to masked regions in images are naturally filled from other frames. The scene template and the warps are jointly inferred. This ensures warps are consistent with the scene and each other as the formulation implicitly imposes that pairwise mappings $w_j\circ w_i^{-1}$ through the scene template must correctly match unmasked portions of $I_i$ and $I_j$. The part of the scene template in view at time $t$ is mapped to a given video frame via the inferred warp to obtain the inpainted result in frame $t$.}
  \label{fig:schematic}
\end{figure*}

We formulate a joint inference problem for the scene template and a set of transformations (warps) of the template to each of the images. The scene template, i.e., the \emph{background}, is a 2D representation of radiance of the scene without the \emph{foreground}, i.e., part of the scene corresponding to masked regions in images to be inpainted. The inference problem arises from a generative model that explains how images are formed through geometric transformations of the scene template; these transformations model transformations arising from both camera viewpoint change and dynamic scenes. This inference constrains motion behind masks to be plausible by being consistent with the generative model and scene template, and hence across video frames. This alleviates problems with frame-to-frame flow propagation approaches, which aim to hallucinate the motion in an image behind the masks without scene consistency. Such a hallucination can lead to errors, which is further amplified over multiple frames through temporal propagation. Given the scene template and transformations, the inpainted result is the mapping of the template to the mask of the image to be inpainted (see Figure~\ref{fig:schematic}).

\cut{and hence across multiple frames from which the template is constructed}

\cut{
and do not carefully handle the occlusion phenomena due to portions of the scene corresponding to one image not being visible in other images, especially over far away image frames.
}

\subsection{Notation and Assumptions}
The video, a set of frames, is denoted $\{I_i\}_{i=1}^{T}$ where $I_i : D \subset \R^2 \to \R^k$ ($k=3$ for RGB values) is an image, $D$ is the image domain, and $T$ is the number of frames. We denote the radiance function of the background as $f : \Omega \to \R^k$, and  $\Omega \subset \R^2$ is the domain of the scene template (often larger than $D$ to accommodate data from all images). We denote the mappings (warps) from the domain of the scene template to each image domain as $\{w_i: \Omega \rightarrow D\}_{i=1}^T$. Note that $w_i$ actually only maps the visible portion of the scene $\Omega$ at frame $i$ to $D$, which is important to deal with moving cameras (details in next section). $w_i$'s are non-rigid, so can handle dynamic scenes/backgrounds. Our model assumes images (outside the mask) are obtained from the scene as $I_i(x) = f(w_i^{-1}(x)) + \eta_i(x)$, where $\eta_i(x)$ is a noise process (to model un-modeled nuisances, e.g., small illumination change, shadows, etc) and $w^{-1}_i$ is the inverse mapping from image $i$ to the template. For the purpose of computation, $\eta$ will be assumed to be a zero-mean Gaussian noise process independent of $x$ and $i$ following Lambertian assumptions.

In video inpainting, masks $M_i$'s are given for each frame. These can either be provided by user annotation or one can leverage object segmentation algorithms. $M_i$ can contain multiple objects (that may move in different ways) of arbitrary size and shape. Inpainting is to retrieve the radiance in the scene behind $M_i$'s.


\subsection{Energy Minimization Formulation}

We now formulate the inference of the scene template $f$ and the warps $w_i$'s as a joint energy minimization problem. Note that if $f$ is known, $w_i$'s can be determined by an optical flow problem. Vice-versa, if $w_i$'s are known, then template radiance can be determined by back warping the region outside the masks to $\Omega$. As neither of them \cut{these variables }are known, the problem is setup as a joint energy minimization on warps and the scene template as follows:
\begin{equation}\label{eq:E_f}
\begin{aligned} 
&E_{f}(f, \{w_i\}_{i=1}^T) = \\
&\sum_{i=1}^T\int_{D\backslash M_i} |I_i(x) - f(w_i^{-1}(x))|_2^2\ud x + \sum_{i=1}^T E_{Reg}(w_i).
\end{aligned} 
\end{equation}
The first term above favors warps (and templates) such that the mapping of the visible part of the scene radiance into the image domain matches the image intensity for all pixels outside the mask $M_i$. Each pixel in each image outside $M_i$ maps to the scene template, and so each $x\in D\backslash M_i$ corresponds to some point in $\Omega$, though not each point in $\Omega$ will correspond to some point in $D$. This is desired since the scene encompasses more than just what is visible from a single frame $I_i$. This is particularly important as we assume that the camera may translate, and so only a portion of the scene template will be visible in frame $I_i$ (see Figure~\ref{fig:schematic}), and thus the first term only penalizes the visible portion of the radiance $f(w_i^{-1}(x)), x\in D\backslash M_i$ in deviating from the image intensity. The second term is warp regularity that is required to make the problem well-posed in light of the aperture problem. We will discuss the particular form of regularization in Section \ref{sec:optimization}.

One can recognize that this formulation is similar to optical flow, but for some key differences. Rather than mapping between frames, we map between images and the scene template (which is to be determined), providing natural consistency of the mappings with the scene and hence also each other, which is not present in previous inpainting works.

\subsection{Optimization}\label{sec:optimization}

\cut{ 
In practice, since the warps are only computed in the background, we adopt SobolevFlow \cite{yang2014shape} due to its region based property. It computes flow tailored to $D\backslash R_i$'s only, and extends to $R_i$'s by spatial regularity. For faster computation, we also offer a deep learning based option by using FlowNet2 \cite{ilg2017flownet} to extract flow in the whole image, and then replace the flow inside by making use of a learned module from \cite{xu2019deep}. 
}

To optimize, we iteratively update the scene template given the current estimate of the warps and vice-versa the warps given the estimate of the template.

\textbf{Update for the scene template:} 
Given estimates of $w_i$'s, $\Omega$ is computed as the union of the back-warpings of each image domain, i.e., $\Omega =\cup_{i=1}^{T}w_i^{-1}(D)$. Note that $\Omega$ can be larger than $D$. We now update the scene template radiance $f$ given $\Omega$. Since $f$ only appears in the first term of \eqref{eq:E_f}, we can ignore the second term to determine $f$. Performing a change of variables to compute the integrals on $\Omega$, allows the summation to be moved inside the integrand. One can then show that the global optimizer for $f$ is:
\begin{equation}\label{eq:f_appearance}
  f^*(p)  = \frac{ \sum_{i=1}^T I_i(w_i(p))\mathbbm{1}_i(w_i(p)) J_i(p)
  }{ \sum_{i=1}^T \mathbbm{1}_i(w_i(p)) J_i(p)}, \quad p\in \Omega
\end{equation}
where $\mathbbm{1}_i(\cdot)$ is the indicator function of $D\backslash M_i$, i.e., $1$ in $D\backslash M_i$, the background region in $I_i$, and $0$ otherwise, and $J_i(p) = \det\nabla w_i(p)$, which results from the change of variables and measures the area distortion in the warp between $\Omega$ and $D$. To obtain the radiance at $p$, one computes a weighted average of all image intensity values from pixels over frames $i$ that correspond to $p$.

\textbf{Update for the warps:} Given an estimate of the template $f^*$, we minimize \eqref{eq:E_f} with respect to $w_i$'s. This is equivalent to computing
\begin{equation}\label{eq:flow}
w_i^*=\argmin_{w_i}\int_{D\backslash M_i} |I_i(x) - f^*(w_i^{-1}(x))|_2^2\ud x + E_{Reg}(w_i),
\end{equation}
for each $i$, which is similar to an optical flow problem, but only the non-mask region ($D\backslash M_i$) is matched, and the shapes of the domains of the $I_i$ ($D$) and $f$ ($\Omega$) are different. For convenience, $w_i$ is extended to the whole domain $\Omega$ by a smooth extension outside the portion of the template visible in frame $i$ through spatial regularity on all of $\Omega$, which is determined by the second term. To naturally handle different shaped domains, we use
SobolevFlow \cite{yang2014shape} to refine the warps initialized by current estimates of $w_i$'s. Our initialization for $w_i$ (described below) uses frame-to-frame flow composition, and the flow update 
\eqref{eq:flow} mitigates errors from the frame-to-frame flow and its aggregation, as it induces geometric consistency (the $w_i$'s must be consistent with the scene template, which is a function of $w_i$'s, and hence geometrically consistent with each other).

\cut{
accumulated by flow composition, allowing aggregating large number of frames without amplifying early flow errors, which is a major advantage over the frame-to-frame scheme. 
}

\begin{algorithm}[t]
\small
  \begin{algorithmic}[1]
    \State Choose a key frame $I_k$
    \State Initialize for $w_i$'s by $w_i = w_{ki}$
    \Repeat{ // {\it updating warp and template} }
    \State $\Omega =\cup_{i=1}^{T}w_i^{-1}(D)$ and compute $f$ by \eqref{eq:f_appearance}
    \State For all $i$, update $w_i$ and $w_i^{-1}$ by computing optical flow between $f$ to $I_i$ // {\it computed in parallel}
    \Until{converges}
  \end{algorithmic}
  \caption{\sl Optimization for the scene template.}
  \label{alg:full}
\end{algorithm}

\textbf{Initialization:} We choose a key frame (e.g., the middle frame) $I_k$ to be the scene template and initialize the warps $w_i$ to be the optical flow between the key frame and frame $i$, i.e., $w_i=w_{ki}$. Note that $k$ and $i$ may represent distant frames, and thus may involve large displacements, challenging for optical flow methods. Therefore, we first compute warp between adjacent frames, i.e. $w_{i(i+1)}$ and $w_{(i+1)i}$, by ordinary optical flow then $w_{ki}$ can be computed as a recursive composition of appropriate consecutive frame flows.

\cut{
$w_{k(i+1)} = w_{i(i+1)} \circ w_{ki}$ and $w_{k(i-1)} = w_{i(i-1)} \circ w_{ki}$. This makes handling large background motion possible. 

In this way, computed by \eqref{eq:E_f}, the initial scene template will align with $I_k$.
}

To compute the frame-to-frame flow, we use SobolevFlow \cite{yang2014shape}, which naturally allows one to exclude the masked region $M_i$ from computation leading to accurate flow. For faster computation, we initialize this with a deep learning based optical flow (FlowNet2 \cite{ilg2017flownet} to extract flow in the whole image, and then replace the flow inside the mask by spatial regularity). We compute flow in both forward and backward direction, and let $w_{ij}^{-1}=w_{ji}$ for the initialization, so all warps have an inverse.

Algorithm \ref{alg:full} summarizes the optimization pipeline. Empirically, it takes at most 2 iterations to converge.

\subsection{Efficient Updates of the Scene Template}\label{sec:faster}
\begin{algorithm}[t]
\small
  \begin{algorithmic}[1]
    \State Initialization: $t=1$, initialize $w_1$ by identity map
    \State $t \leftarrow t + 1$, a new frame $I_t$ and mask $R_t$ available \label{step:1}
    \State Compute $w_{t,t-1}$ and $w_{t,t-1}^{-1}$ by optical flow
    \State For all $i$'s, update warps by $w_i \leftarrow w_{t(t-1)}\circ w_{i}$
    \Repeat{ // {\it updating warp and template} }
    \State Update template $f$ by \eqref{eq:f_appearance}
    \State For all $i$, update $w_i$'s and $w_i^{-1}$'s by optical flow between $I_i$ and $f$ // {\it computed in parallel}
    \Until{converges}
    \State Compute inpainting result by \eqref{eq:I_appearance}
    \State Go to Step \ref{step:1}
  \end{algorithmic}
  \caption{\sl Scene template / Inpainting  (faster).}
  \label{alg:faster}
\end{algorithm}

When processing longer videos with large range of background motion, $\Omega$ can grow arbitrarily large, reducing computational and memory efficiency. However, one usually does not need a full scene template to inpaint a frame $I_i$, since $w^{-1}_{i}(M_{i})$ usually only maps to a small portion of the template. Therefore, we further propose an efficient implementation of our approach. To do this, we ensure that the scene template is aligned to the newest frame as follows. Suppose the template aligns with $I_t$, then given a new frame $I_{t+1}$, we update the warps to align to $I_{t+1}$ through compositions $w_i \leftarrow w_{i}\circ w_{(t+1)t}$ and $w_i^{-1} \leftarrow w_{t(t+1)}\circ w_{i}^{-1}$ for all $i$'s, and $f$ is updated by \eqref{eq:f_appearance}. We can then crop $\Omega$ to be $D$ (matching $I_{t+1}$), and $w^{-1}_{t+1}(M_{t+1})$ is likely to remain in $D$ since $w^{-1}_{t+1}$ is close to the identity map, as the initial template aligns with $I_{t+1}$. This also makes refinement of $w_i$'s easier to be handled by existing optical flow methods, as the domains of the $I_i$ and $f$ are the same. 

Further, $w_i$'s are computed in parallel since the update of each are independent. Algorithm \ref{alg:faster} shows this scheme, which is used for the experiments. Since each frame is only inpainted by previous frames, there may be holes remaining in the initial frames. To alleviate this, we do a forward sweep to inpaint every frame followed by a backward sweep, which fills the initial frames. In the experiments, We find using a sliding window of 7 frames to solve the optimization already provides good results. Nevertheless, our method can handle much larger number of frames (e.g. 60 frames in Figure \ref{fig:schematic}) without hitting hardware constraints.

\section{Inpainting}
\def\figd{Figures/L1L2}
\def\fWidD{0.12\textwidth}
\begin{figure}[t]
\centering
{\scriptsize
\hspace{-0.3in}
\begin{tabular}[0mm]{c@{\hskip 0.01in}c@{\hskip 0.01in}c@{\hskip 0.01in}c@{\hskip 0.01in}c@{\hskip 0.01in}c@{\hskip 0.01in}c}

\rotatebox{90}{\quad Images}
\includegraphics[width=\fWidD]{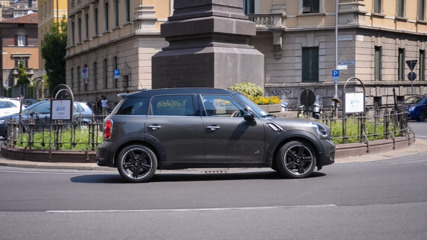} &
\includegraphics[width=\fWidD]{\figd/org_40.jpg} &
\includegraphics[width=\fWidD]{\figd/org_50.jpg} &
\includegraphics[width=\fWidD]{\figd/org_60.jpg} &
\\
\rotatebox{90}{\quad\quad L2}
\includegraphics[width=\fWidD]{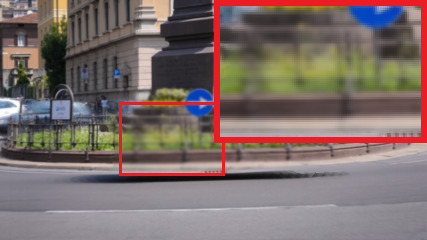} &
\includegraphics[width=\fWidD]{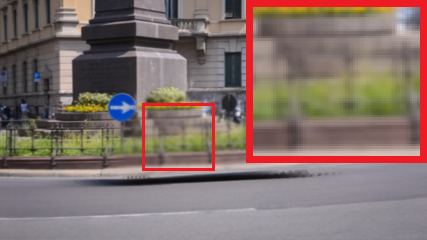} &
\includegraphics[width=\fWidD]{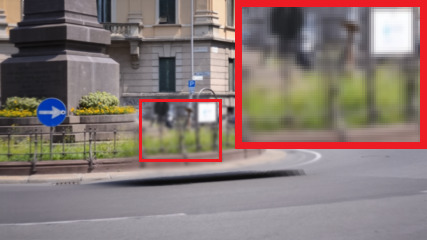} &
\includegraphics[width=\fWidD]{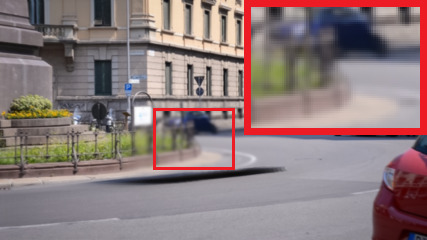} &
\\
\rotatebox{90}{\,\,\, L2+L1}
\includegraphics[width=\fWidD]{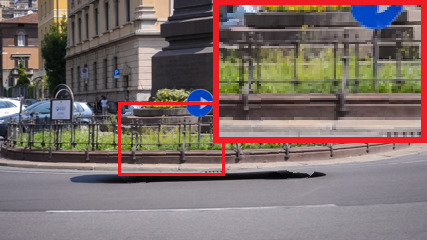} &
\includegraphics[width=\fWidD]{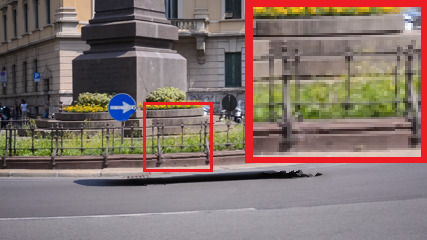} &
\includegraphics[width=\fWidD]{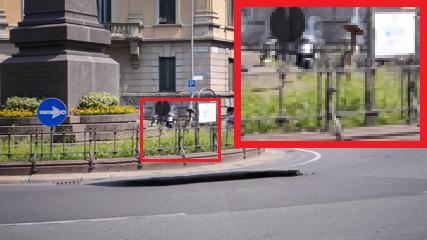} &
\includegraphics[width=\fWidD]{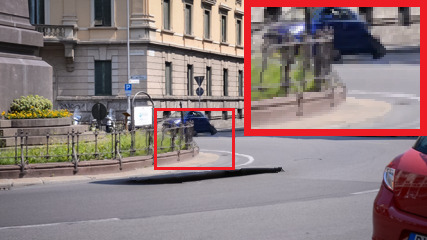} &
\end{tabular}
}
\caption{\small {\bf Example: $L^2$ vs $L^2+L^1$.} $L^2$ creates smooth but blurry inpainting results. By $L^1$ regularization, the inpainting result preserves sharp and rigid appearance.}
\label{fig:L1L2}
\end{figure}

Although we can simply map the template $f$ into the masked region $M_i$ to be inpainted via the warp $w_i$ computed in the previous section to produce the inpainting, this can result in blurry results, as the $L^2$ norm in \eqref{eq:E_f} can result in a blurry template as a result of temporal averaging in \eqref{eq:f_appearance}. To mitigate these effects, we solve an energy minimization problem for the inpainted radiance $P_t : M_t \to \R^k$ that interpolates between the template $f$ and raw image values from other frames that map to the mask, through the $L^1$ distance that reduces temporal blurring, as follows:
\begin{equation}\label{eq:E_I}
\begin{aligned}
E_{img}(P_t) = &\int_{M_t}\left[ |P_t(x) - f^*(w_t^{-1}(x))|_2^2 \right. \\&+ \beta\sum_{i=1}^n \left.|P_t(x) - I_i(w_{ti}(x))|_1   \mathbbm{1}_t(w_{ti}(x)) \right] \ud x,
\end{aligned}
\end{equation}
where the first term measures fidelity of the inpainting to the template mapped into $M_t$, and the second measures fidelity of the inpainting to mappings of other frames into $M_t$. $\mathbbm{1}_t(w_{ti}(x))$ indicates whether $w_{ti}(x)$ is visible in frame $i$.

Let $\{t_x^1, \dots, t_x^{m_x}\}$ denote the $m_x$ frames where $w_{ti}(x)$ maps into $M_t$. From \cite{li2009new}, \eqref{eq:E_I} has a closed form minimizer:
\begin{equation}\label{eq:I_appearance}
\hspace{-0.15in}
\begin{aligned} 
P^*_t(x) = \median \{& I_{t_x^1}(w_{tt_x^1}(x)), \cdots, I_{t_x^{m_x}}(w_{tt_x^{m_x}}(x)),\\f^*(y) - \frac{m_x}{2}\beta, &f^*(y) - (\frac{m_x}{2}+1)\beta , \cdots, f^*(y) + (\frac{m_x}{2})\beta \},
\end{aligned}
\end{equation}
where $y=w^{-1}_t(x)$. The inpainting is a {\it temporal} median filtering of pixels from different frames that map into the mask as well as the template appearance. Figure \ref{fig:L1L2} shows an example of the reduced blurriness of this $L^1$ formulation.

\begin{figure}[t]
\scriptsize
\centering 
\def\fWidD{0.18\textwidth}
\begin{subfigure}[t]{0.45\textwidth}
\begin{tabular}[0mm]{c@{\hskip 0in}c@{\hskip 0in}c@{\hskip 0in}c@{\hskip 0in}c}
Template&Flow&GT&Bilinear & N. Neighbor
\\
\includegraphics[width=\fWidD]{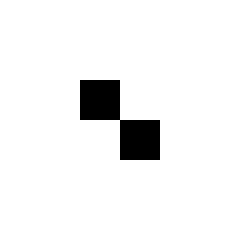} &
\includegraphics[width=\fWidD]{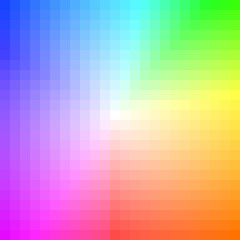} &
\includegraphics[width=\fWidD]{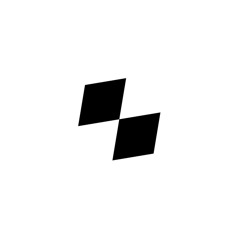}&
\includegraphics[width=\fWidD]{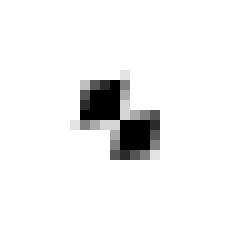}&
\includegraphics[width=\fWidD]{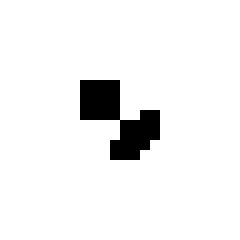}
\end{tabular}
\caption{\sl\small A toy template propagated by a rotational flow.}\label{fig:interp}
\end{subfigure}
\hfill
\def\fWidD{0.34\textwidth}
\begin{subfigure}[t]{0.47\textwidth}
\hspace{-0.15in}
\begin{tabular}[0mm]{c@{\hskip 0.01in}c@{\hskip 0.01in}c}
Bilinear & N.Neighbor & Our Best Practice
\\
\includegraphics[width=\fWidD]{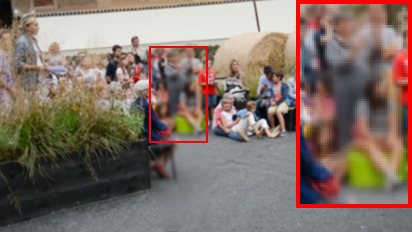}&
\includegraphics[width=\fWidD]{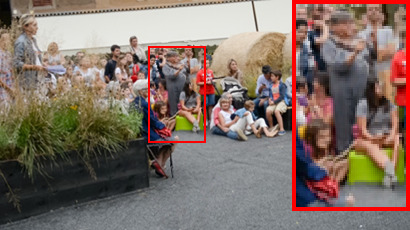} &
\includegraphics[width=\fWidD]{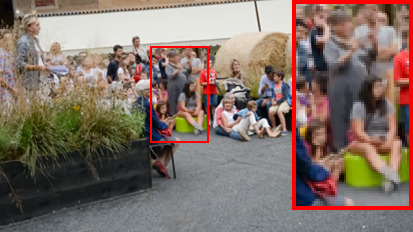}
\end{tabular}
\caption{\sl\small Inpainting results: different interpolation methods.}\label{fig:interp2}
\end{subfigure}
\caption{\small {\bf Interpolation artifacts.} Bilinear interpolation induces blurring; nearest-neighbor interpolation creates distortion. Our combination achieves the most realistic result.}
\end{figure}

We now discuss the evaluation of the images at transformed pixels in \eqref{eq:I_appearance}, which requires interpolation. This choice is key to producing a visually plausible result. Common choices are nearest neighbors or bilinear interpolation. In Figure \ref{fig:interp}, we propagate a toy template by a rotational optical flow and show the effects of each scheme. Bilinear preserves the shape but edges are blurred since the pixel value is a linear combination of white and black pixels around the endpoint. Nearest-neighbor preserves pixel values and thus is not blurred. However, rounding endpoints to nearest pixels lead to geometric distortion. Defects from both interpolation schemes are further amplified and propagated throughout the whole video in flow guided inpainting.

\cut{
In inpainting, image values need to be propagated across many frames and this blurring becomes noticable.
}

We propose a simple yet effective solution to this long-ignored problem in flow-based inpainting that our framework naturally suggests. We apply bilinear interpolation in computing the template \eqref{eq:f_appearance} to avoid geometric distortion. After obtaining a smooth template with well-aligned warps, nearest-neighbor interpolation is applied in computing the inpainting in \eqref{eq:I_appearance} to preserve rigid edges. This achieves the result with the least artifacts. As suggested by Figure \ref{fig:interp2}, only using bilinear interpolation leads to a blurry result while only using nearest-neighbor interpolation leads to distortion. The best result is obtained by our combination.

\def\figd{Figures/interp}

After the inpainting described above, there may be some masked pixels that are not filled, as they correspond to points in the scene that never revealed in the entire video. To fill these pixels, we use DeepFill \cite{yu2018generative}, following \cite{xu2019deep,Gao-ECCV-FGVC}.

\begin{table*}[h!]
\centering
{
\small
\begin{tabular}{|c|cc|cc|cc|ccc|}
\hline
 &\multicolumn{4}{c|}{DAVIS}&\multicolumn{5}{c|}{Foreground Removal}\\\cline{2-10}
&TPSNR$\uparrow$&TSSIM$\uparrow$&Avg$\uparrow$&\#1$\uparrow$&TPSNR$\uparrow$&TSSIM$\uparrow$&PSNR$\uparrow$&SSIM$\uparrow$&FID$\downarrow$\\
\hline
Ours+F+S&29.08&0.881&-&-&\bf{34.58}&\bf{0.935}&\bf{29.15}&\bf{0.857}&\bf{1.042}\\
Ours+S&\bf{30.86}&\bf{0.922}&{\bf7.24}&{\bf581}&30.14&0.895&28.17&0.825&1.179\\
DFG\cite{xu2019deep}&27.58&0.852&6.89&457&30.74&0.886&28.28&0.803&1.376\\
ILA\cite{zhang2019internal}&30.59&0.894&5.12&95&31.35&0.910&25.64&0.769&1.292\\
OnionPeel\cite{oh2019onion}&26.24&0.817&-&-&29.31&0.822&28.71&0.833&1.051\\
STTN\cite{zeng2020learning}&29.04&0.874&-&-&34.07&0.926&27.62&0.817&1.136\\
DeepFill\cite{yu2018generative}&19.56&0.554&3.96&24&20.77&0.639&19.31&0.568&2.546\\
\hline
FGVC \cite{Gao-ECCV-FGVC}&30.71&0.916&-&-&32.89&0.936&31.76&0.886&0.833\\
Ours+FGVC flow &\bf{30.94}&\bf{0.921}&-&-&\bf{34.56}&\bf{0.947}&\bf{31.89}&\bf{0.894}&\bf{0.802}\\
\hline
\end{tabular}
}
\vspace{-0.08in}\caption{\small {\bf Quantitative Results}. On both datasets, our method achieves the best performance in terms of temporal consistency and inpainting quality. Our results are more preferred in the user study. F: FlowNet2 \cite{ilg2017flownet}; S: SobolevFlow \cite{yang2014shape}.}
\vspace{-0.1in}
\label{tab:comparison}
\end{table*}

\section{Experiments}
To the best of our knowledge, currently, there is no video inpainting benchmark dataset, but only for object segmentation (DAVIS \cite{Perazzi2016}). While it is possible to use the segmented masks from DAVIS for video inpainting (e.g. \cite{xu2019deep,Gao-ECCV-FGVC}), there is no ground truth and evaluation relies on user studies.

As a complement, many methods \cite{xu2019deep,zhang2019internal,kim2019deep,zeng2020learning} compose moving objects or masks over background videos, so inpainting accuracy can be numerically evaluated. Since data are created in different ways (some are not publicly released), a direct quantitative comparison is infeasible. Therefore, we introduce a new dataset under this setting, called {\it Foreground Removal} dataset, with quantitative evaluation protocols measuring inpainting accuracy and temporal consistency. The dataset will be made publicly available.

{\bf DAVIS} \cite{Perazzi2016} contains a total of 3455 frames in 50 videos with pixel-wise per-frame annotations. The task is to remove annotated moving objects. We perform a user study by inviting 24 volunteers from both inside and outside the field to rate the inpainting quality of each video from 1 to 10. Videos are displayed at 15 fps and users can stop, replay, and zoom-in freely. The ordering of the methods is randomly permuted (not known to the users). Each user is required to rate at least 15 sequences and we collected 871 results in total. We also evaluate temporal consistency following \cite{zhang2019internal} by measuring visual similarity of the inpainted region in adjacent frames, labeled as TPSNR and TSSIM.

{\bf Foreground Removal dataset} includes 25 composed videos ranging from 29 to 90 frames whose backgrounds are collected from Youtube. We paste moving foreground from DAVIS and SegtrackV2 \cite{TsaiBMVC10} to background videos. The dataset contains representative challenging cases including viewpoint change, deforming background, illumination change, fast zooming-in, motion blur, image noise (e.g. rain), and constant (e.g. overexposed) regions. We evaluate inpainting accuracy by computing PSNR, SSIM, and Fréchet Inception Distance \cite{heusel2017gans} to the ground truth, and evaluate temporal consistency as in DAVIS.

\def\figd{Figures/user_study}
\begin{figure}[t]
\centering
\hspace{-0.2in}
\begin{tabular}[0mm]{c@{\hskip 0.01in}c}
\includegraphics[height=1.15in]{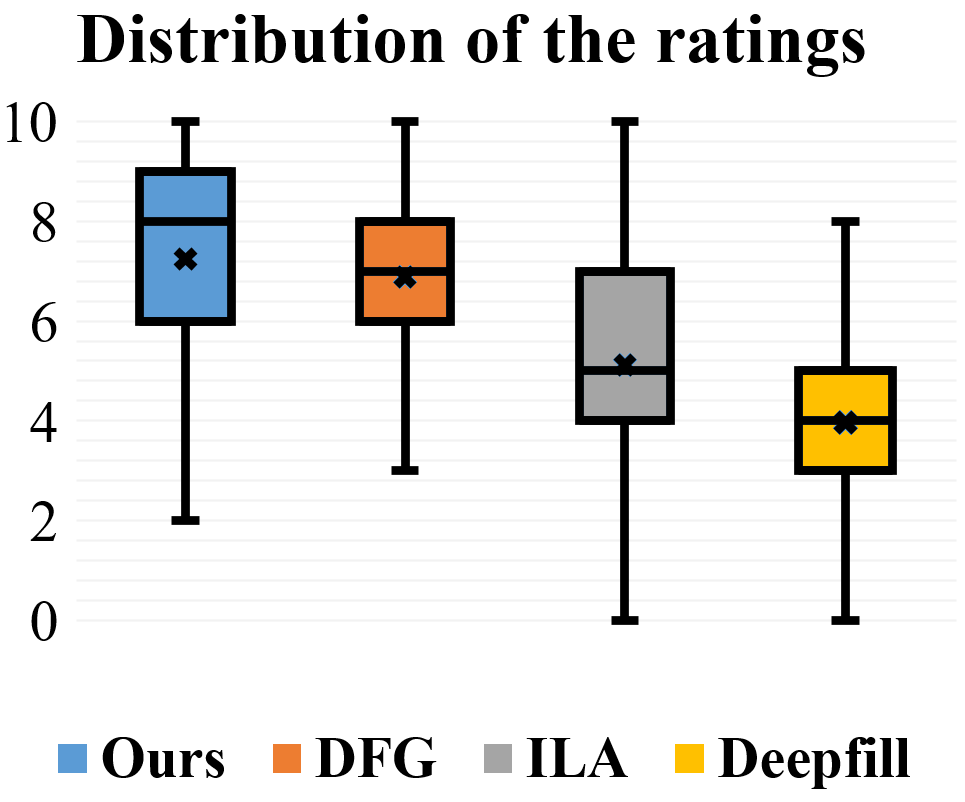} &
\includegraphics[height=1.15in]{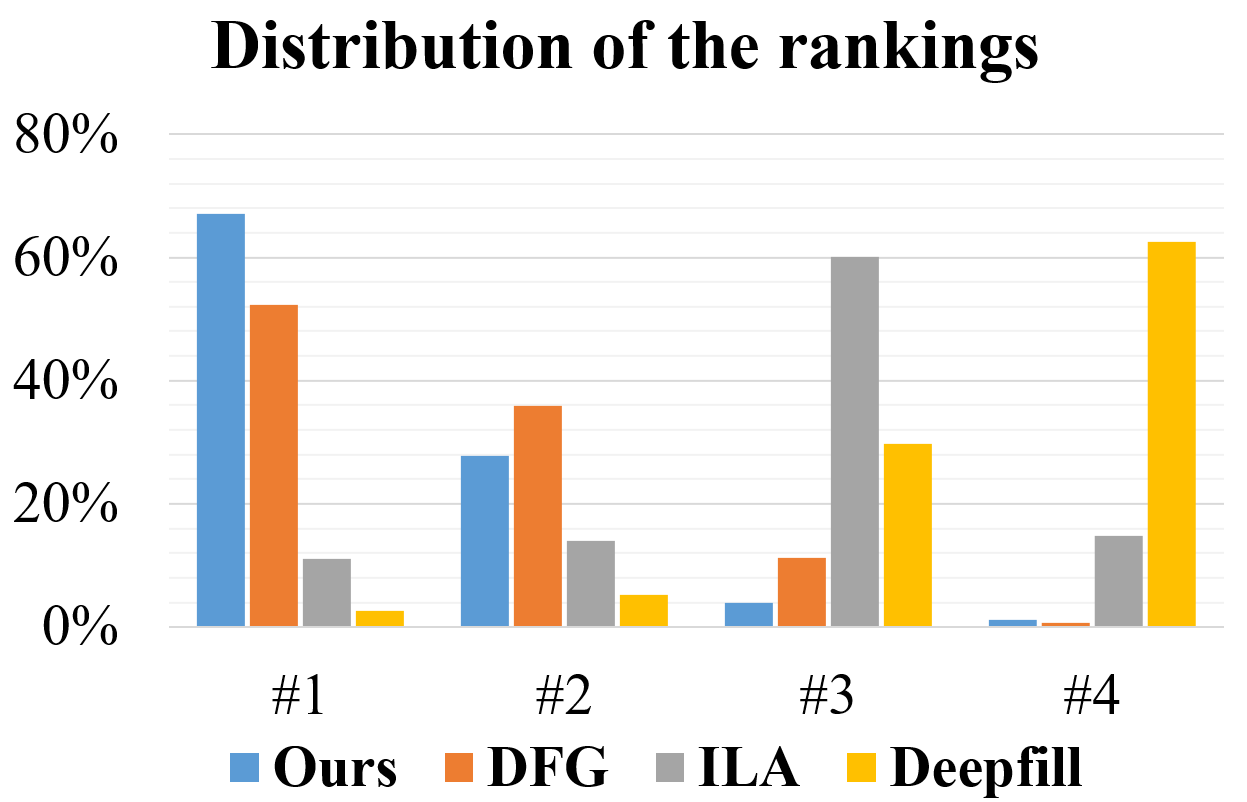}
\end{tabular}
\vspace{-0.1in}
\caption{\small {\bf User study on DAVIS.} Ours achieves highest average and median rating. Ours also receives the most \# 1 ranking.}
\label{fig:user}
\vspace{-0.1in}
\end{figure}

{\bf Comparison:} We compare our approach with state-of-the-art methods. They are: flow-guided {\it DFG} \cite{xu2019deep} and {\it FGVC} \cite{Gao-ECCV-FGVC}; end-to-end {\it ILA} \cite{zhang2019internal}, {\it OnionPeel} \cite{oh2019onion}, {\it STTN} \cite{zeng2020learning}; and single image {\it DeepFill} \cite{yu2018generative}. For the user study, we choose DFG, ILA and DeepFill since they are representative of each category of method. Due to author-released code of \cite{zhang2019internal,xu2019deep,oh2019onion} operating at different resolutions, we resize all results to the same resolution for a fair numerical and visual comparison. As described in Section \ref{sec:optimization}, we apply SobolevFlow \cite{yang2014shape} for flow refinement ({\it Ours+S}). To be comparable to \cite{xu2019deep}, we use FlowNet2 \cite{ilg2017flownet} for flow initialization ({\it Our+F+S}). We compare to FGVC \cite{Gao-ECCV-FGVC}, which uses a more advanced flow method. For this comparison, we initialize our method with the flow used by \cite{Gao-ECCV-FGVC}.

\subsection{Results}
\def\figd{Figures/comparison}
\def\fWidD{0.16\textwidth}
\begin{figure*}[t]
\centering
{\scriptsize
\begin{tabular}[0mm]{c@{\hskip 0.01in}c@{\hskip 0.01in}c@{\hskip 0.01in}c@{\hskip 0.01in}c@{\hskip 0.01in}c@{\hskip 0.01in}c@{\hskip 0.01in}c}
\rotatebox{90}{\quad Image}
\includegraphics[width=\fWidD]{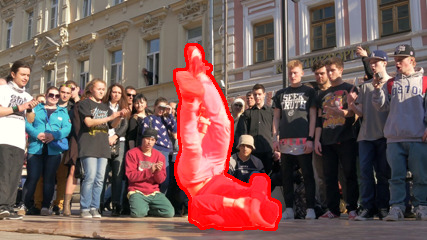} &
\includegraphics[width=\fWidD]{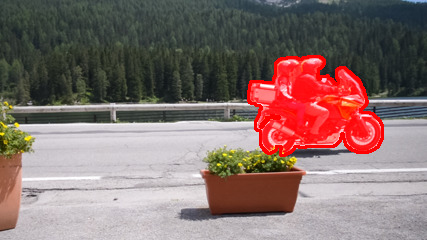} &
\includegraphics[width=\fWidD]{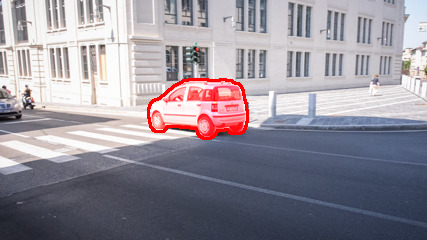} &
\includegraphics[width=\fWidD]{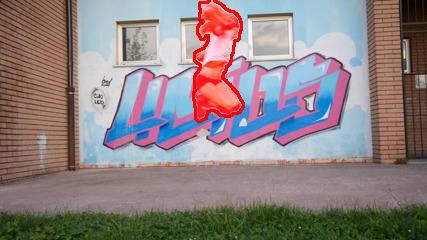} &
\includegraphics[width=\fWidD]{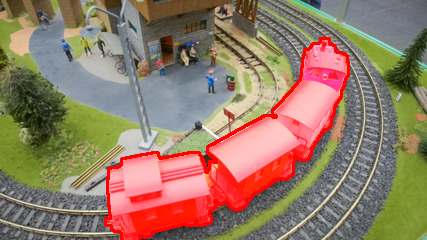}&
\includegraphics[width=\fWidD]{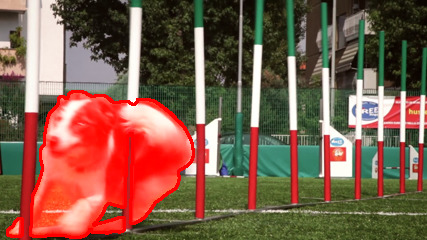}
\\
\rotatebox{90}{\,\, Ours+S}
\includegraphics[width=\fWidD]{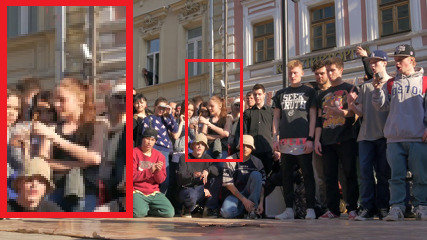} &
\includegraphics[width=\fWidD]{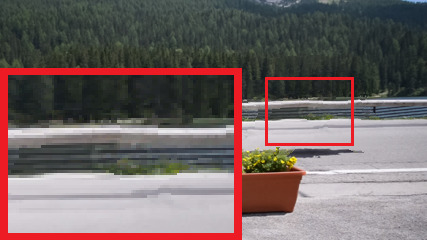} &
\includegraphics[width=\fWidD]{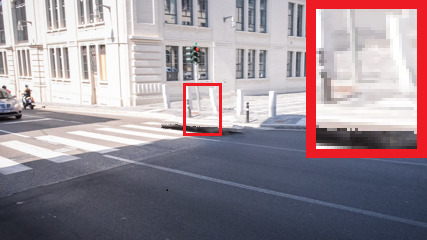} &
\includegraphics[width=\fWidD]{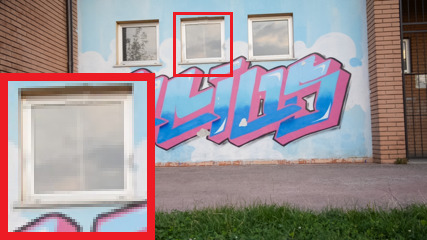} &
\includegraphics[width=\fWidD]{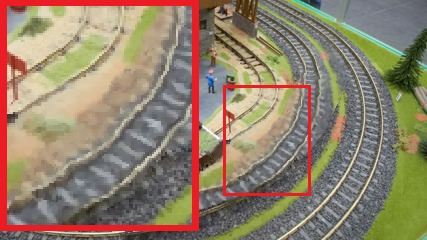}&
\includegraphics[width=\fWidD]{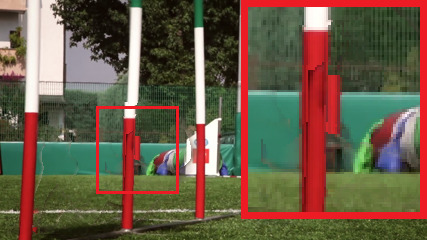} &
\\
\rotatebox{90}{Ours+F+S}
\includegraphics[width=\fWidD]{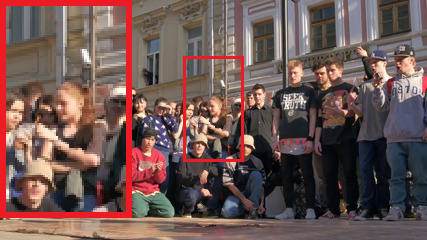} &
\includegraphics[width=\fWidD]{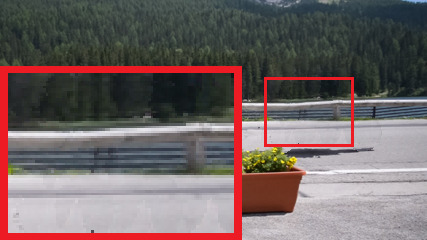} &
\includegraphics[width=\fWidD]{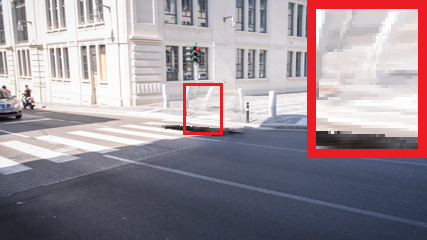} &
\includegraphics[width=\fWidD]{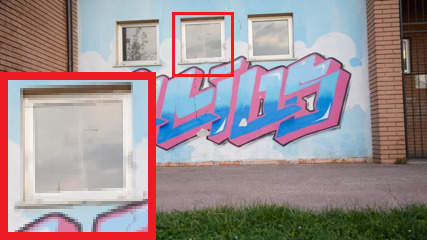} &
\includegraphics[width=\fWidD]{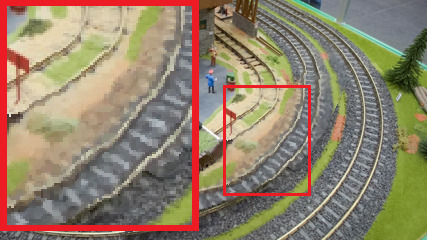}&
\includegraphics[width=\fWidD]{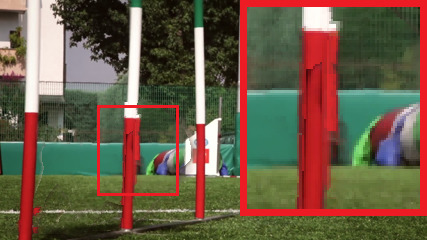} &
\\
\rotatebox{90}{\quad\quad DFG}
\includegraphics[width=\fWidD]{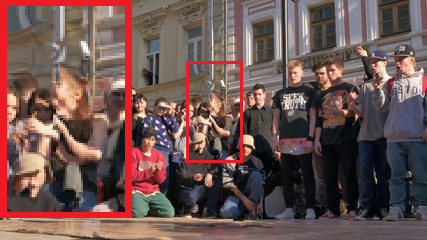} &
\includegraphics[width=\fWidD]{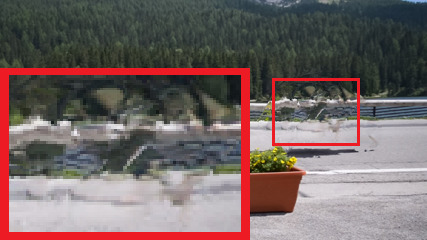} &
\includegraphics[width=\fWidD]{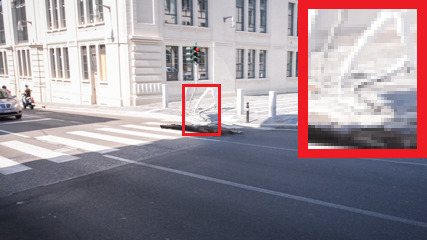} &
\includegraphics[width=\fWidD]{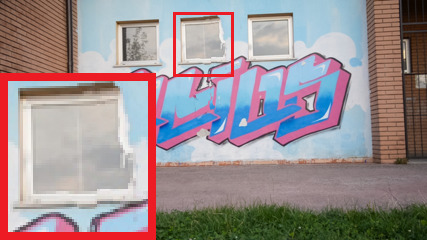} &
\includegraphics[width=\fWidD]{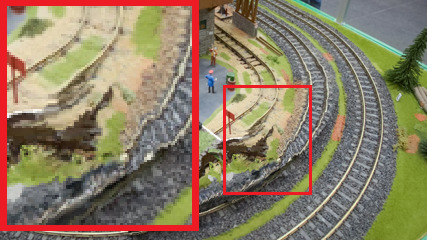} &
\includegraphics[width=\fWidD]{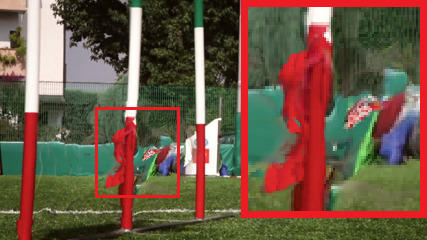}

\\
\rotatebox{90}{\quad OnionPeel}
\includegraphics[width=\fWidD]{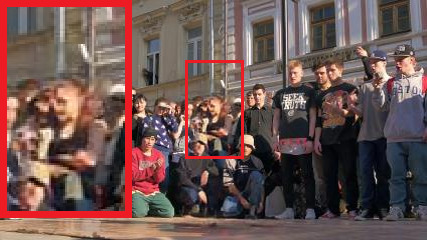} &
\includegraphics[width=\fWidD]{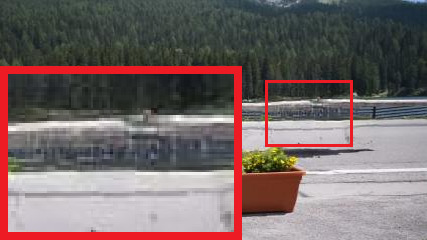} &
\includegraphics[width=\fWidD]{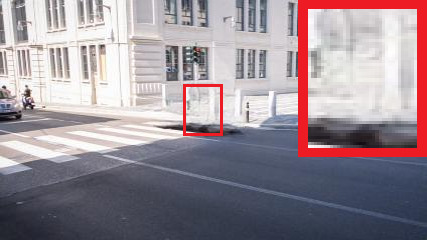} &
\includegraphics[width=\fWidD]{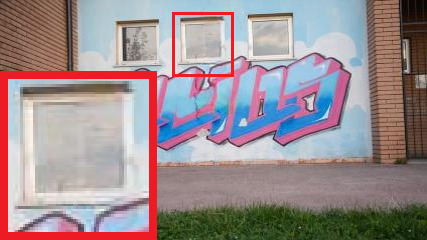} &
\includegraphics[width=\fWidD]{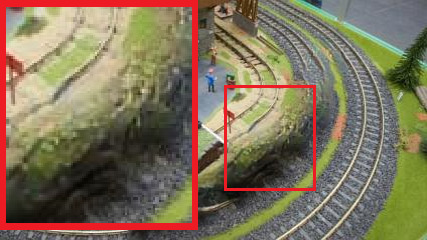} &
\includegraphics[width=\fWidD]{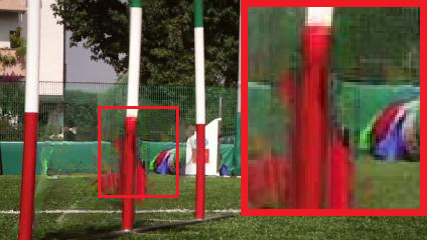}
\\
\rotatebox{90}{\quad \quad ILA}
\includegraphics[width=\fWidD]{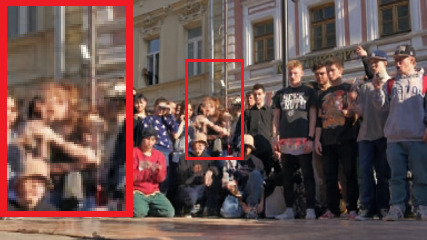} &
\includegraphics[width=\fWidD]{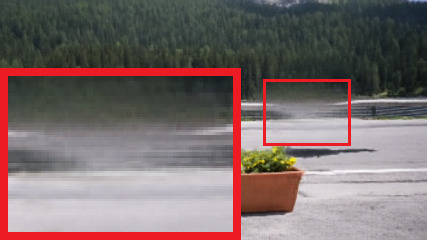}  &
\includegraphics[width=\fWidD]{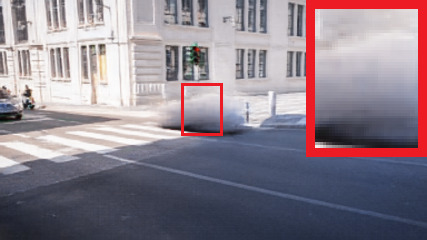} &
\includegraphics[width=\fWidD]{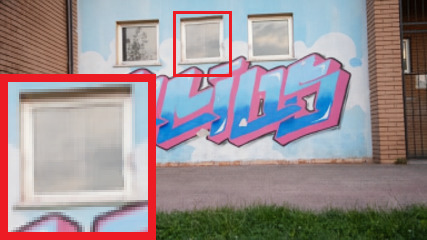} &
\includegraphics[width=\fWidD]{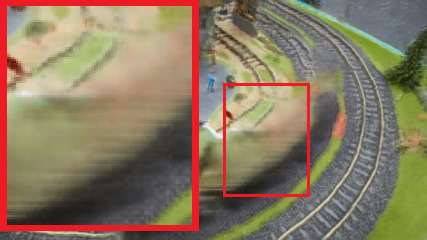}&
\includegraphics[width=\fWidD]{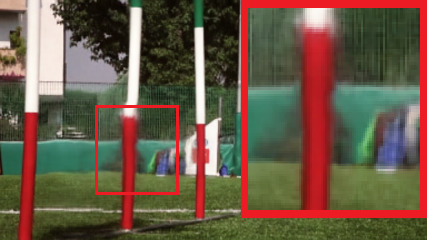}
\end{tabular}
}
\caption{\small {\bf Comparison on the DAVIS dataset.} With our scene template and corresponding optimization and interpolation scheme, our method obtains the most realistic results. {\it Better viewed zoomed in.}}
\label{fig:comparison1}
\vspace{-0.1in}
\end{figure*}

\def\figd{Figures/rebuttal/watermark}
\def\fWidD{0.115\textwidth}
\begin{figure}[t]
\centering
{\scriptsize
\hspace{-0.27in}
\begin{tabular}[0mm]{c@{\hskip 0.01in}c@{\hskip 0.01in}c@{\hskip 0.01in}c@{\hskip 0.01in}c@{\hskip 0.01in}c@{\hskip 0.01in}c}
\rotatebox{90}{\quad Images}
\includegraphics[width=\fWidD]{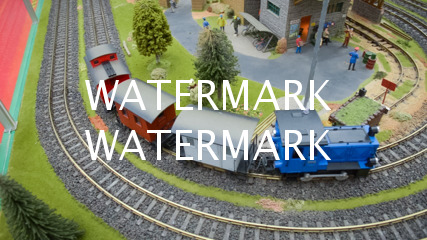} &
\includegraphics[width=\fWidD]{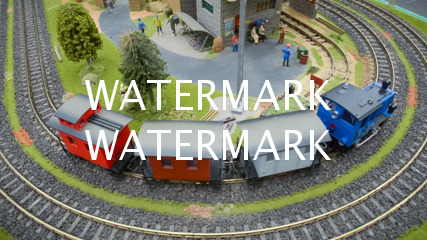} &
\includegraphics[width=\fWidD]{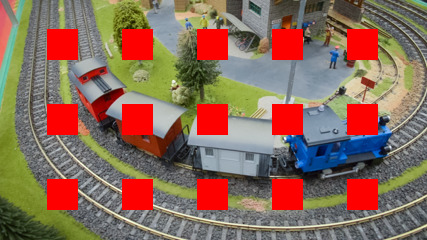} &
\includegraphics[width=\fWidD]{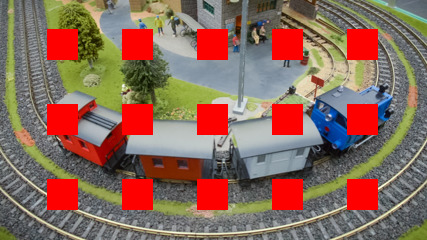} &
\\
\rotatebox{90}{\quad Results}
\includegraphics[width=\fWidD]{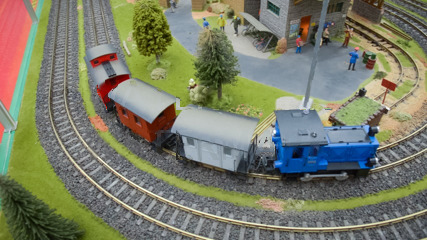} &
\includegraphics[width=\fWidD]{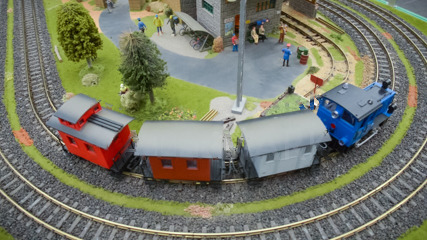} &
\includegraphics[width=\fWidD]{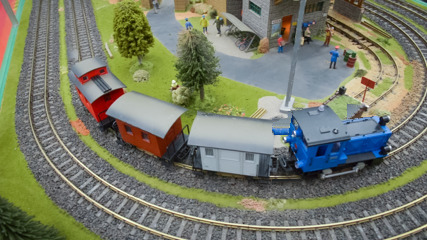} &
\includegraphics[width=\fWidD]{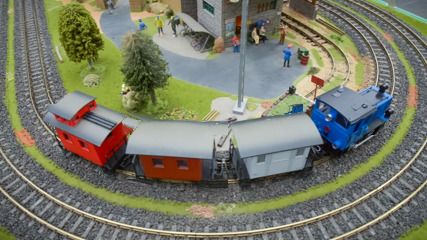} &
\end{tabular}
}
\vspace{-0.1in}
\caption{\small {\bf Sample fixed region removal from a dynamic scene (from DAVIS)}. Animation in the supplementary materials.}
\vspace{-0.1in}
\label{fig:fixed}
\end{figure}

\def\fWidD{0.24\textwidth}
\begin{figure}[t]
{\scriptsize
\begin{subfigure}[t]{0.5\textwidth}
\hspace{-0.3in}
\def\figd{Figures/annotation}
\begin{tabular}[0mm]{c@{\hskip 0.01in}c@{\hskip 0.01in}c@{\hskip 0.01in}c@{\hskip 0.01in}c}
&\multicolumn{4}{c}{\includegraphics[width=0.94\textwidth]{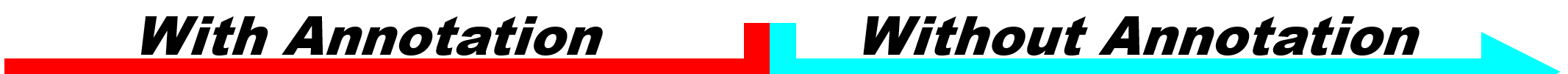}}\\
\rotatebox{90}{\,\,Images}&
\includegraphics[width=\fWidD]{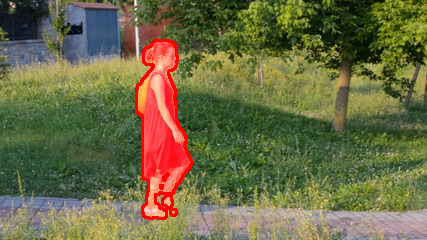}&
\includegraphics[width=\fWidD]{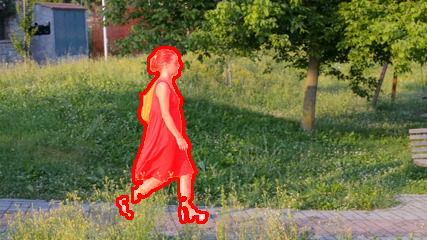}&
\includegraphics[width=\fWidD]{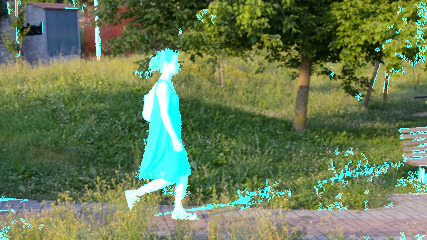}&
\includegraphics[width=\fWidD]{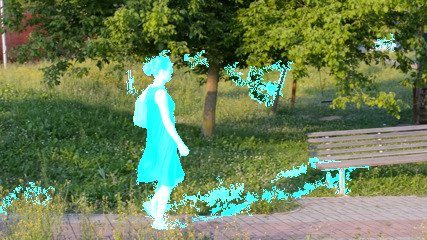} 
\\
\rotatebox{90}{\,\,Results}&
\includegraphics[width=\fWidD]{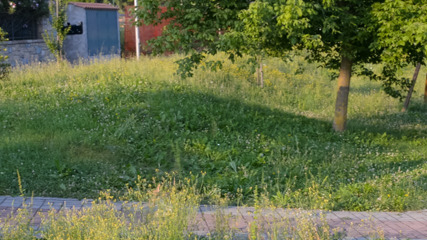}&
\includegraphics[width=\fWidD]{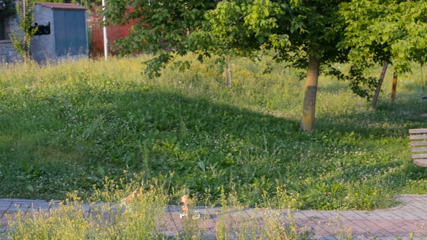}&
\includegraphics[width=\fWidD]{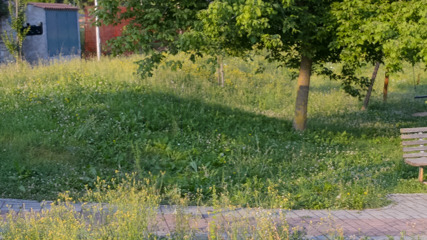}&
\includegraphics[width=\fWidD]{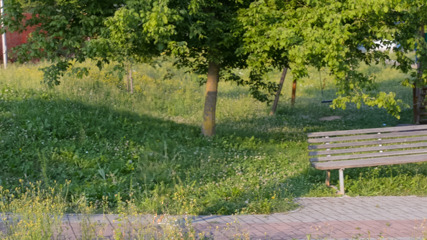}
\end{tabular}
\caption{\sl\small Inpainting with missing annotations.}
\end{subfigure}

\begin{subfigure}[t]{0.5\textwidth}
\hspace{-0.2in}
\def\figd{Figures/shadow}
\begin{tabular}[0mm]{c@{\hskip 0.01in}c@{\hskip 0.01in}c@{\hskip 0.01in}c}
Labels&Estimated&Original Result&Shadow removal  
\\
\includegraphics[width=\fWidD]{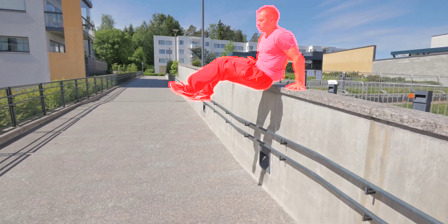}&
\includegraphics[width=\fWidD]{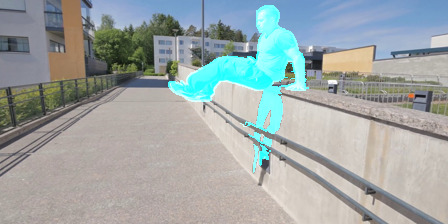}&
\includegraphics[width=\fWidD]{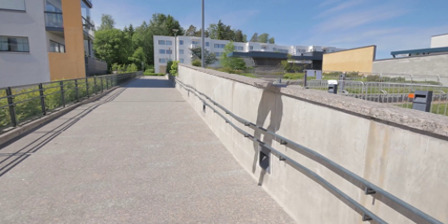}&
\includegraphics[width=\fWidD]{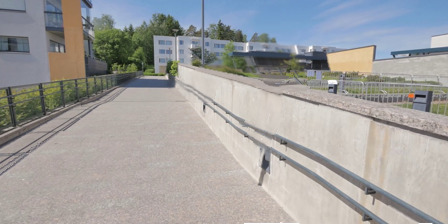} 
\\
\includegraphics[width=\fWidD]{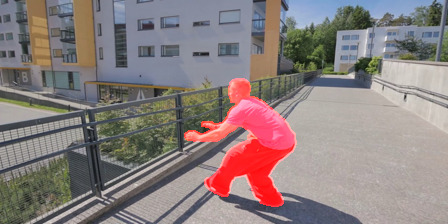}&
\includegraphics[width=\fWidD]{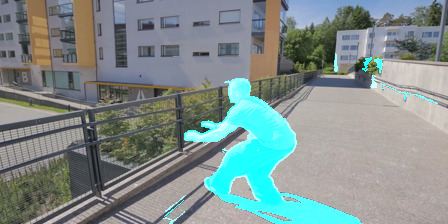}&
\includegraphics[width=\fWidD]{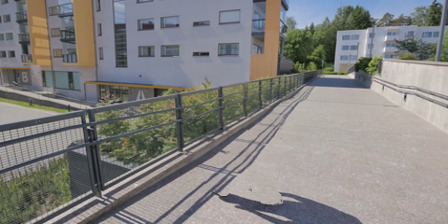}&
\includegraphics[width=\fWidD]{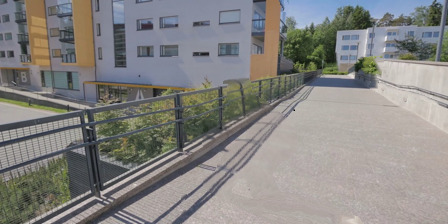} 
\end{tabular}
\caption{\sl\small Automatic shadow estimation and removal.}
\end{subfigure}
}
\vspace{-0.1in}
\caption{\small {\bf Inpainting with incomplete annotations.} Our method can be applied to correct incomplete annotations. It estimates missing annotations by \eqref{eq:est_r} and performs inpainting. {\it Red: given annotations; blue: estimated masks.}}
\label{fig:annotation}
\vspace{-0.1in}
\end{figure}

\def\figd{Figures/comparison2}
\def\fWidD{0.16\textwidth}
\begin{figure*}[t]
\centering
{\scriptsize
\begin{tabular}[0mm]{c@{\hskip 0.01in}c@{\hskip 0.01in}c@{\hskip 0.01in}c@{\hskip 0.01in}c@{\hskip 0.01in}c@{\hskip 0.01in}c@{\hskip 0.01in}c}
\rotatebox{90}{\quad Image}
\includegraphics[width=\fWidD]{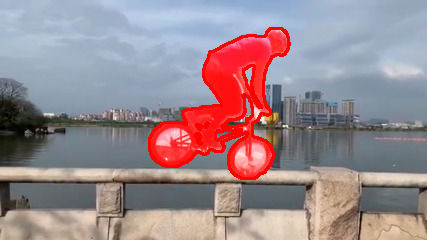} &
\includegraphics[width=\fWidD]{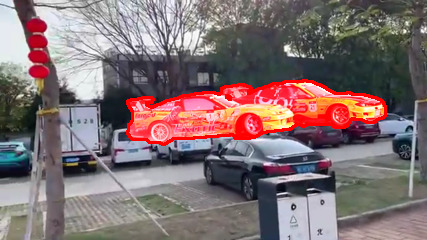} &
\includegraphics[width=\fWidD]{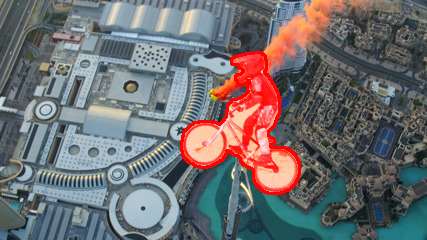} &
\includegraphics[width=\fWidD]{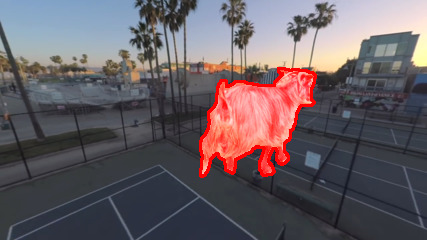} &
\includegraphics[width=\fWidD]{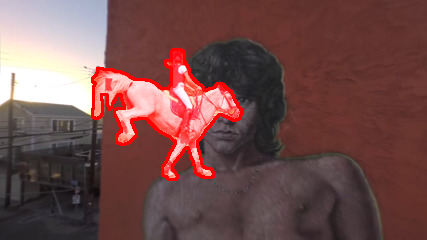} &
\includegraphics[width=\fWidD]{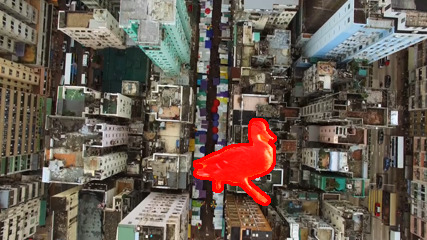}
\\
\rotatebox{90}{\quad\quad GT}
\includegraphics[width=\fWidD]{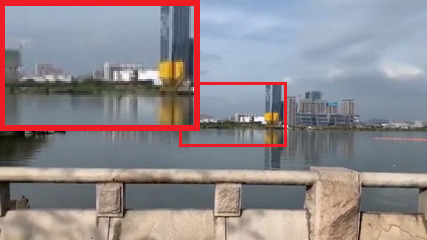} &
\includegraphics[width=\fWidD]{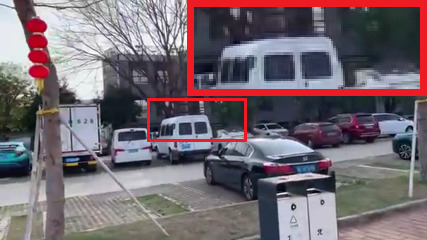} &
\includegraphics[width=\fWidD]{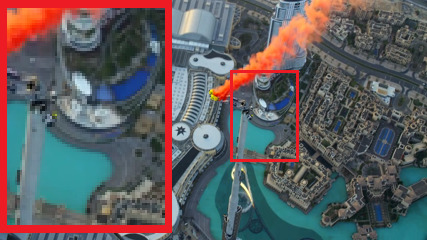} &
\includegraphics[width=\fWidD]{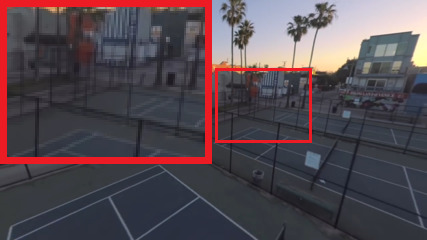} &
\includegraphics[width=\fWidD]{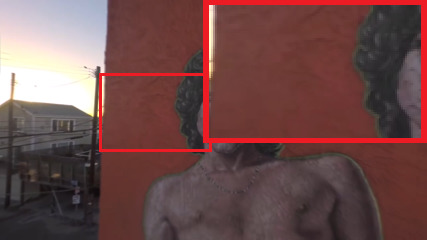} &
\includegraphics[width=\fWidD]{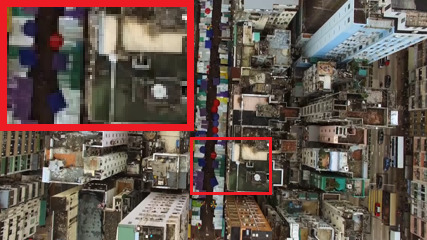}
\\
\rotatebox{90}{Ours+F+S}
\includegraphics[width=\fWidD]{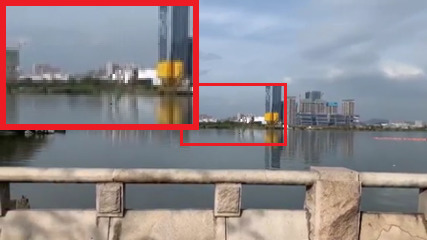} &
\includegraphics[width=\fWidD]{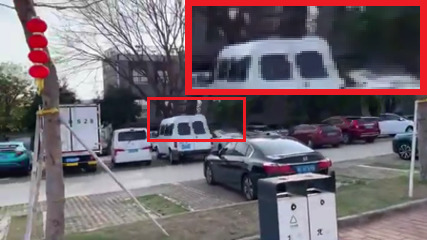} &
\includegraphics[width=\fWidD]{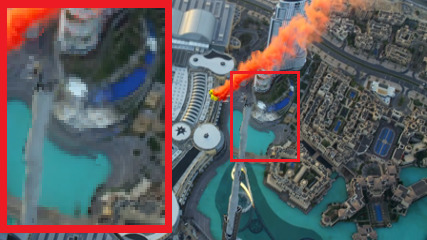} &
\includegraphics[width=\fWidD]{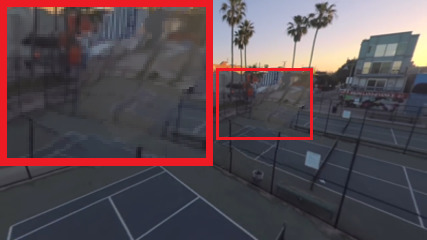} &
\includegraphics[width=\fWidD]{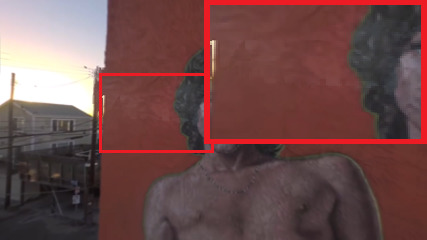} &
\includegraphics[width=\fWidD]{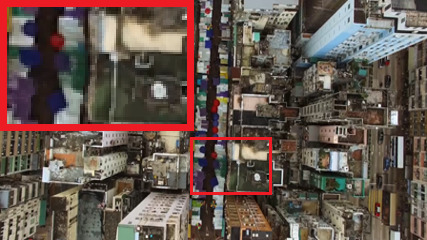}
\\
\rotatebox{90}{\quad\quad DFG}
\includegraphics[width=\fWidD]{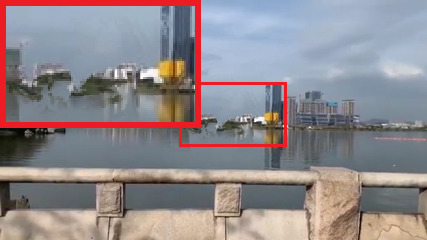} &
\includegraphics[width=\fWidD]{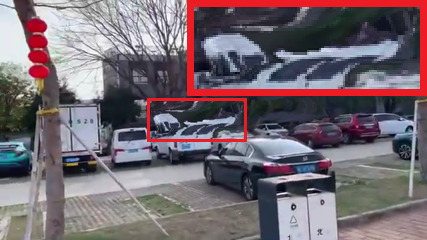} &
\includegraphics[width=\fWidD]{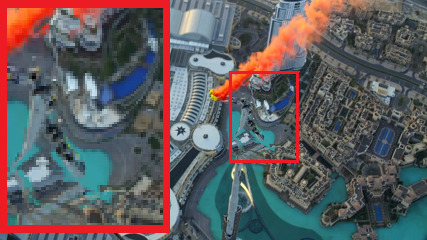} &
\includegraphics[width=\fWidD]{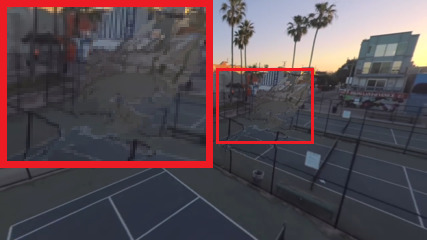} &
\includegraphics[width=\fWidD]{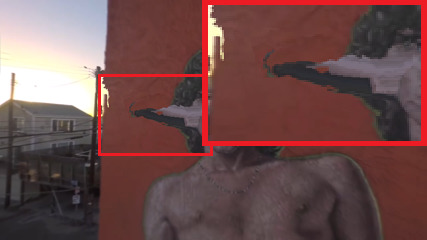} &
\includegraphics[width=\fWidD]{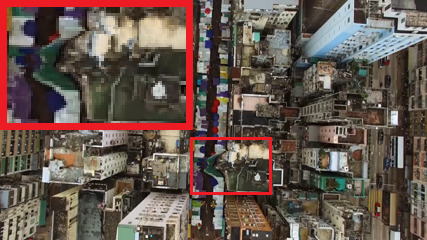}
\\
\rotatebox{90}{\quad OnionPeel}
\includegraphics[width=\fWidD]{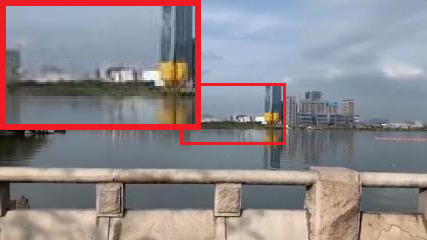} &
\includegraphics[width=\fWidD]{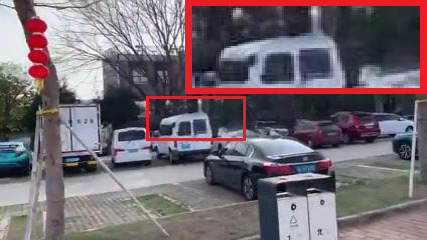} &
\includegraphics[width=\fWidD]{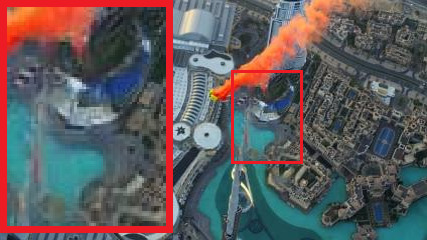} &
\includegraphics[width=\fWidD]{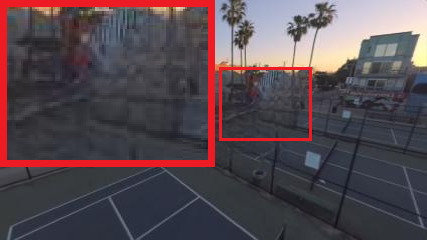} &
\includegraphics[width=\fWidD]{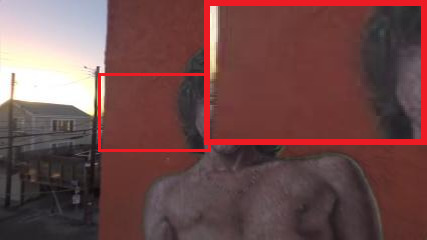} &
\includegraphics[width=\fWidD]{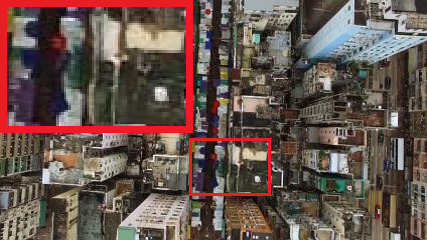}
\\
\rotatebox{90}{\quad \quad ILA}
\includegraphics[width=\fWidD]{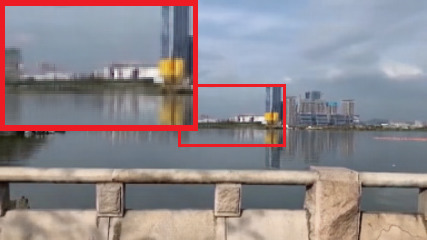} &
\includegraphics[width=\fWidD]{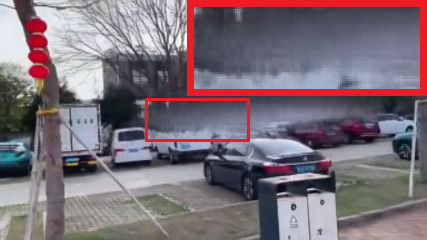} &
\includegraphics[width=\fWidD]{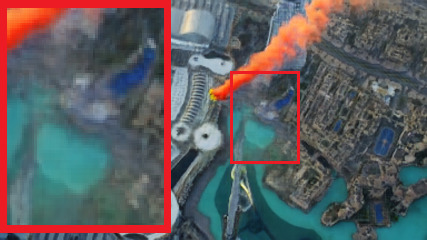} &
\includegraphics[width=\fWidD]{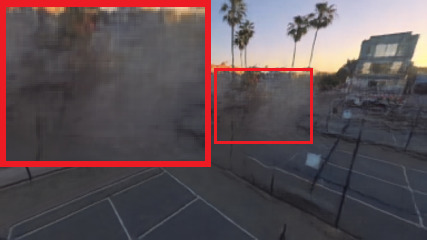} &
\includegraphics[width=\fWidD]{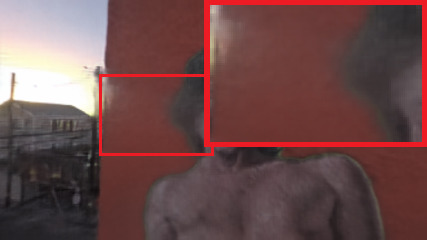} &
\includegraphics[width=\fWidD]{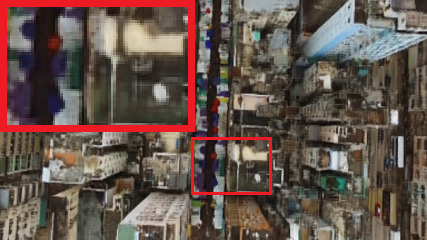}
\end{tabular}
}
\caption{\small {\bf Comparison on Foreground Removal dataset.} Our novel dataset contains challenging cases including irregular background motion, illumination change, fast zooming-in, motion blur, image noise, and constant regions. Our method obtains the most visually plausible and temporally consistent results. {\it Better viewed zoomed in.}}
\label{fig:comparison2}
\end{figure*}

\textbf{DAVIS Dataset}: Figure \ref{fig:comparison1} shows representative visual results on DAVIS. Ours are more visually plausible than competing methods. DFG is vulnerable to distortion due to frame-to-frame propagation and nearest-neighbor interpolation. OnionPeel and ILA have blurry results (frequently observed in other learning-based methods \cite{zeng2020learning,kim2019deep,lee2019copy}, too). Our method preserves rigid object appearances, showing the effectiveness of the $L^2$-$L^1$ optimization and interpolation strategy. Ours also significantly outperforms the competition in videos with long-term occlusion (e.g. Figure \ref{fig:teaser}) since the method is less vulnerable to flow error accumulation.

Figure \ref{fig:user} summarizes the user study. Our method has the best average user rating. It also receives the highest count of number one rankings (tie allowed). In 67\% of the ratings, ours rank the best among the four methods. Table \ref{tab:comparison} shows quantitative results. {\it Ours+S} achieves the best temporal consistency. This is presumably because the region-based formulation of SobolevFlow provides a more consistent background motion estimation. Initialized by the same flow as \cite{Gao-ECCV-FGVC}, we achieve better temporal consistency, which shows that the scene consistency in our method improves over even more advanced optical flow (also see Figure \ref{fig:teaser}).

\textbf{Foreground Removal Dataset}: Figure \ref{fig:comparison2} and Table \ref{tab:comparison} show qualitative and quantitative results on the dataset. Ours obtains visually plausible results. Since the dataset contains more challenging background motion and FlowNet2 has a stronger capability to handle complex motion, Ours+F+S achieves dominant performance. Similar to DAVIS, our method improves over \cite{Gao-ECCV-FGVC}.

We want to highlight two cases in Figure \ref{fig:comparison2}: in the car scene (2nd column), our approach successfully handles two foreground objects with different motions; in the horse scene (5th column) our approach handles strong camera zooming, a challenging case to video inpainting.

\cut{
\begin{table}[b!]
\centering
{
\small
\begin{tabular}{|c|cc|cc|cc|}
\hline
&\multicolumn{2}{c|}{DAVIS}&\multicolumn{4}{c|}{Foreground Removal}\\
\cline{2-7}&\!\!TPSNR\!\!\!\!&TSSIM\!\!&\!\!TPSNR\!\!\!\!&TSSIM\!\!&\!\!PSNR\!\!\!\!&SSIM\!\!\\
\hline
\cite{Gao-ECCV-FGVC}&30.71&0.916&32.89&0.936&31.76&0.886\\
\!\!+ Ours\!\!&\bf{30.94}&\bf{0.935}&\bf{34.56}&\bf{0.947}&\bf{31.89}&\bf{0.894}\\
\hline
\end{tabular}
}
\vspace{-0.08in}\caption{\small {\bf Quantitative Comparison with \cite{Gao-ECCV-FGVC}}. Adopting the same optical flow, Our method improves over \cite{Gao-ECCV-FGVC} on all metrics.}
\vspace{-0.1in}
\label{tab:comparison2}
\end{table}}

{\bf Fixed Regions (DAVIS)}: The literature considers removal of content within a fixed region of video in scenes with dynamic objects. Our formulation \eqref{eq:E_f} can, in principle, be applied to this case (Figure \ref{fig:fixed}). However, for high accuracy, one requires additional occlusion reasoning in optical flow, which will be subject of future work; the focus of the current paper is to illustrate the benefits of the scene template. Even with our current methodology, our results are comparable to \cite{Gao-ECCV-FGVC}, the state-of-the-art for fixed regions: PSNR 28.02 vs 28.20, SSIM 0.959 vs 0.957 on DAVIS following the experimental setups of \cite{Gao-ECCV-FGVC}. We provide detailed discussions in the supplementary materials.

\subsection{Further Application and Discussion}
\textbf{Incomplete annotations}: One task that has not been considered by previous work is handling incomplete or inperfect annotations. Existing methods (e.g. \cite{huang2016temporally,xu2019deep,zhang2019internal,Gao-ECCV-FGVC}) assume perfect annotations. In practice, masks often come from user annotation or segmentation algorithms, and so annotations may not be available for the whole video or may contain errors. Figure \ref{fig:annotation} shows two examples. In the first, masks from only the first 10 frames are provided; in the second, shadows are not included in masks. Our method can estimate foreground masks by thresholding the residual between the scene template (computed from available noisy annotations) and images:
\begin{equation}\label{eq:est_r}
R_t = \{|I_t(x)-f(w_t^{-1}(x))|_2^2 > \alpha\}, \quad \alpha = 0.1.
\end{equation}
Our method infers missing annotations and corrects incorrect annotations, and so to the best of our knowledge, the first to inpaint despite inperfect annotations. This makes fully automated foreground removal algorithm possible, which can be a future direction of research.

\def\figd{Figures/failure_hack}
\begin{figure}
\vspace{-0.1in}
\centering
\small
\def\fWidD{0.12\textwidth}
\!\!\!\!\!\!\!\!
\begin{tabular}[0mm]{c@{\hskip 0.01in}c@{\hskip 0.01in}c}
Image&DFG&DFG+Ours\\
\includegraphics[width=\fWidD]{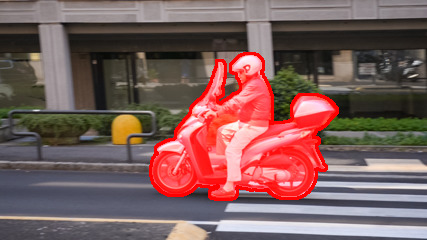} &
\includegraphics[width=\fWidD]{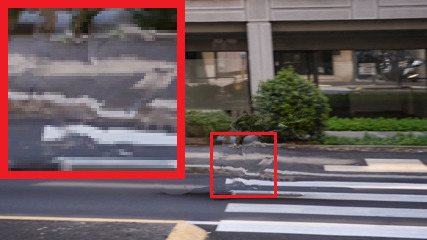} &
\includegraphics[width=\fWidD]{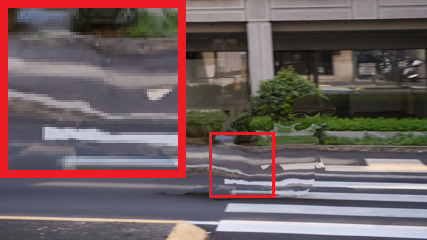}\\
\includegraphics[width=\fWidD]{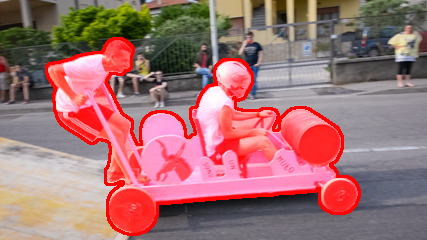} &
\includegraphics[width=\fWidD]{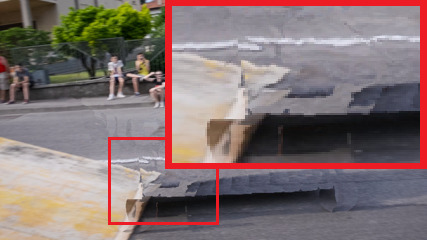} &
\includegraphics[width=\fWidD]{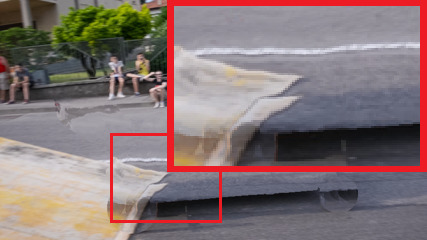}
\end{tabular}
\caption{\small {\bf Multi-frame aggregation improves regularity} Even if using the same optical flow, our method shows stronger regularity than frame-to-frame propagation.}
\label{fig:discussion}
\end{figure}
\textbf{Multi-frame aggregation improves regularity}: In this experiment, we use the same optical flow as DFG (without refinement), so the only difference is multi-frame aggregation v.s. frame-to-frame propagation. On DAVIS, TPSNR rises from 27.58 to 30.53 and TSSIM rises from 0.852 to 0.966, showing stronger temporal consistency. We can even observe results with stronger spatial regularity, shown in Figure \ref{fig:discussion}. This shows the advantage of our scene template.

\cut{
\textbf{Speed}: Our approach (Algorithm~\ref{alg:faster}) runs at 3 secs per frame on DAVIS, which is comparable to state-of-the-art flow-guided inpainting methods \cite{xu2019deep,Gao-ECCV-FGVC}. The bottleneck is optical flow, which we expect to improve.
}

\section{Conclusion}
We proposed a novel method for flow-guided video inpainting by introducing the \emph{scene template}, which is a 2D representation of the background. The method aggregates appearance information across frames into the scene template by non-rigid maps, which are solved jointly, then maps the template to the images for inpainting. This results in more plausible and temporally consistent flows than existing flow-based methods as the maps must be consistent with the \emph{scene}. We proposed a simple interpolation scheme, which significantly reduced inpainting artifacts. Experiments showed that our method achieved state-of-the-art results on two datasets in terms of inpainting accuracy and temporal consistency. Our method can also handle missing and noisy user mask annotations.

\clearpage
\bibliographystyle{ieee}
\bibliography{main}
\end{document}